\documentclass[letterpaper, 10 pt, conference]{ieeeconf}  

\IEEEoverridecommandlockouts                              

\usepackage{graphicx} 
\usepackage{times} 
\usepackage{amsmath} 
\usepackage{amssymb}  

\graphicspath{{./figure/}}

\title{\LARGE \bf
Tracking Human-like Natural Motion Using Deep Recurrent Neural Networks     
}

\author{Youngbin Park, Sungphill Moon and Il Hong Suh}

\begin{document}

\maketitle
\thispagestyle{empty}
\pagestyle{empty}

\begin{abstract}
Kinect skeleton tracker is able to achieve considerable human body tracking performance in convenient and a low-cost manner. However, The tracker often captures unnatural human poses such as discontinuous and vibrated motions when self-occlusions occur. A majority of approaches tackle this problem by using multiple Kinect sensors in a workspace. Combination of the measurements from different sensors is then conducted in Kalman filter framework or optimization problem is formulated for sensor fusion. However, these methods usually require heuristics to measure reliability of measurements observed from each Kinect sensor. In this paper, we developed a method to improve Kinect skeleton using single Kinect sensor, in which supervised learning technique was employed to correct unnatural tracking motions. Specifically, deep recurrent neural networks were used for improving joint positions and velocities of Kinect skeleton, and three methods were proposed to integrate the refined positions and velocities for further enhancement. Moreover, we suggested a novel measure to evaluate naturalness of captured motions. We evaluated the proposed approach by comparison with the ground truth obtained using a commercial optical maker-based motion capture system.


\end{abstract}

\section{Introduction}

The second version of the device, the Microsoft Kinect for Window v2(Kinect v2), was released and made available to researchers in 2014. This new generation of Kinect sensor offers a higher resolution and a wider field of view compared to the original Kinect technology. Further, in terms of depth, Kinect v2 is based on time-of-flight principle, whereas the previous version of Kinect utilized structured light to reconstruct the third dimension. This difference has led a considerable improvement in the accuracy of depth sensing.

To enable the use of Kinect sensors for developers and researchers, the official Microsoft SDKs (Software Development Kits) 1.0 and 2.0 are freely available for Kinect v1 and v2, respectively. These SDKs provide a set of functions, especially including human body skeleton tracker. Due to the enhanced depth sensor, tracking accuracy has been improved in Kinect V2. Therefore, in this work we developed our skeleton tracking system based on Kinect v2. 

Although Kinect v2 provides better tracking results comparing to Kinect v1, it often captures unnatural skeleton poses such as discontinuous and vibrated motions in the presence of self-occlusion, which is common among most vision-based sensing systems. A simple way to solve this problem is to use multiple cameras in the workspace. For instance, if a view of a body part is blocked from one camera, it might be possible to obtain a view of the body part from another camera. Subsequently, appropriately combining data obtained from multiple Kinect sensors can be used to achieve more accurate tracking compared with a single sensor. A majority of approaches integrate the measurements from different sensors in Kalman filter framework or formulate optimization problem for sensor fusion. However, these methods require way to estimate the confidence of each measurement for combining multiple observations based on the confidence level. This usually leads heuristic measure to evaluate reliability of the measurements.

In this paper, we developed a method to improve Kinect skeleton using single Kinect sensor, in which supervised learning technique was employed to correct unnatural tracking motions. Specifically, deep recurrent neural networks were used for improving joint positions and velocities of Kinect skeleton data, and three methods were proposed to integrate the refined joint positions and velocities for further enhancement. Consequently, the proposed method removes jitters and promotes temporal continuity. Moreover, we suggested a novel measure to evaluate naturalness of captured motions.

The remainder of the paper is organized as follows. Section 2 provides a survey of the current literatures related to the topic of improvement of Kinect skeleton. Section 3 briefly describes how to improve joint positions and velocities of Kinect skeleton data using deep recurrent neural network. In Section 4, three methods are proposed to integrate the enhanced position and velocity. A novel measure to evaluate naturalness of captured motions is given in Section 5. Section 6 presents our experimental setup and evaluation of the performance of the proposed model. Finally, we present out conclusions in Section 7.

\section{Related Works}

Skeleton tracking algorithms can be classified into single-view based models \cite{Park08}, \cite{Ziegler06}, \cite{Hofmann11} and multi-view based model \cite{Baak11}, \cite{Zhang13}. Shotton \emph{el al.} \cite{Shotton11} proposed a new method to predict 3D positions of body joints from a single depth image. In their method, an intermediate representation of body parts was designed to map the pose estimation problem onto a per-pixel classification problem. An extensively large and highly varied training data set is employed for the random forest classifier to estimate body parts invariant to pose, body shape, clothing, etc. Finally, confidence-scored 3D proposals of several body joints are generated by re-projecting the classification results to the 3D world and finding local modes. As a result, this approach can quickly and accurately predict the 3D positions of body joints. The skeleton trackers in both the first and second versions of the Kinect SDK are based on this algorithm. However, the 3D body pose that is estimated using a single view frequently has problems of determining positions of joints during self-occlusion motions. Consequently, Kinect skeleton tracker has problems of capturing discontinuous movements or unwanted vibration. 

Therefore, approaches that utilize multiple views have recently begun to receive significant attention. For example, Zhang \emph{el al.} \cite{Zhang12} fused individual depth images to a joint point cloud and used an efficient particle filtering approach for pose estimation. Likewise, Liu \emph{el al.} \cite{Liu13} presented a markerless motion capture approach for multi-view video that reconstructs the skeletal motion and detailed surface geometries of two closely interacting people. The approach presented in this paper differs from the methods used by studies described above. Specifically, our goal was not to develop a method that estimates 3D positions of body joint directly from raw depth images or RGB images, but rather to investigate how to generate more human-like natural motion by improving the estimated Kinect v2 skeleton.

Indeed, there have been relatively few studies to determine skeleton pose by enhancing Kinect skeleton tracking. Masse \emph{el al.} \cite{Masse13} presented a framework that obtains 3D positions of body joints from multiple Kinect sensors and then inputs the measured skeletons into a Gated Kalman Filter. In their method, the gated Kalman Filter rejects skeleton poses if the measurement residual referred to as \emph{innovation} is lower than the gating threshold. This is done in order to discard faulty sensor readings and retain correct measurements. For quantitative evaluation, commercial motion capture system is used to get access to the ground truth. However, the processing step to reject measurement is quite simple and entirely relies on \emph{innovation}. This might be often possible to lead ineffective measurement fusion.

Yeung \emph{el al.} \cite{Yeung13} developed a method synthesizing skeletons with duplex Kinect sensors that capture human motion in different views. In their study, each joint had two measurements reported by two cameras. The major technical difficulty comes from how to evaluate the reliability of the two values at each joint, and how to resolve any inconsistencies. To address this problem, they developed a measure to estimate confidence on the 3D positions obtained using the Kinect skeleton tracker. Specifically, the distances between a joint $i$ and the closest joint $j$ estimated from Kinect $A$ and the distance between corresponding joint $i$ and the closest joint $k$ estimated from Kinect $B$ are computed, then if the distance between $i$ and $j$ is smaller than the distance between $i$ and $k$ the joint $i$ obtained from Kinect $A$ is considered as unreliable estimation otherwise, the joint $i$ obtained from Kinect $B$ is considered as the mis-leading joint. This reliability was computed in advance before data fusion procedure based on mathematical optimization was executed. Data fusion procedure was formulated under the mathematical optimization problem, in which objective is to reduce sum of differences between the estimated joint position and the corresponding more reliable position, and the bone-lengths are given as equality constraints.

Both studies described above are different to our approach in following two reason: First, single Kinect sensor was used in our method. Second, We formulate our problem as supervised learning task instead of employing simple Kalman filtering or formulating mathematical optimization problem. In terms of these two aspects, an approach similar to our method has not been proposed. 

%

\section{Improving Position and Velocity of Kinect Skeleton using Deep Recurrent Neural Network}

First part of our method is to improve joint position and velocity of Kinect skeleton using supervised learning. The inputs for the supervised learning are sequences of 3D position or velocity obtained by Kinect skeleton tracker and the targets are sequences of skeleton pose captured using commercial optical maker-based motion capture system. In our method, deep recurrent neural network is employed to solve the regression problem, in which two deep recurrent neural networks are trained separately for refining positions and velocities of body joints. In this Section, we will briefly describe deep recurrent neural network and present the detail of how to train the networks.  

\subsection{Deep Recurrent Neural Network}

A recurrent neural network (RNN) \cite{Rumelhart86} is a neural network that simulates a discrete-time dynamical system and are a powerful model for sequential data. A conventional RNN is constructed by defining the transition function and the output function as

\begin{eqnarray}
\label{eq:RNN}
\textbf{h}_{t} &=& \phi_{h}\left( \textbf{W}^{T}\textbf{h}_{t-1}+\textbf{U}^{T}\textbf{x}_{t} \right)  \\
\textbf{y}_{t} &=& \phi_{o}\left(	\textbf{V}^{T}\textbf{h}_{t} \right),
\end{eqnarray}

\noindent where $\phi_{h}$, $\phi_{o}$, $\textbf{x}_{t}$, $\textbf{y}_{t}$ and $\textbf{h}_{t}$ are respectively a state transition function, an output function, an input, an output, a hidden state, and $\textbf{W}$, $\textbf{U}$ and $\textbf{V}$ are the transition, input and output matrices, in that order. It is usual to use a nonlinear function such as a logistic sigmoid function or a hyperbolic tangent function for $\phi_{h}$.

Deep learning is built based on a hypothesis that a deep, hierarchical model can be exponentially more efficient at representing some functions than a shallow one \cite{Bengio09}. Several theoretical results and empirical evidences support this hypothesis \cite{Roux10}, \cite{Goodfellow13}, \cite{Delalleau11}. \noindent RNNs are inherently deep in time, since their hidden state is a function of all previous hidden states. However, the potential weakness for RNNs is that RNNs lack hierarchical processing of the input in space. From this perspective view, deep recurrent neural networks has recently gained significant attention to many researchers. As with feedforward deep neural networks have multiple nonlinear layers between input and output, a recurrent network can be considered as a deep recurrent neural network (DRNNs) if the network has more than one hidden layers. 

We can now consider two schemes of DRNNs. One has $L$ hidden layer with temporal connection only at the $l$-th layer and the other has $L$ hidden layer with full temporal connections (called stacked RNN). Based on empirical evaluation on our datasets, we have chosen the former scheme. The $l$-th hidden activation at time $t$, $\textbf{h}^{l}_{t}$, is defined as

\begin{eqnarray}
\label{eq:RNN}
\textbf{h}^{l}_{t} \!\!\!\!\! &=& \!\!\!\!\! \phi_{l}\left( {\textbf{W}^{l}}^{T} \!\! \textbf{h}_{t-1} \! + \! {\textbf{U}^{l}}^{T} \!\! \phi_{l-1} \! \left( {\textbf{U}^{l-1}}^{T} \!\! \left(   \ldots \phi_{1} \! \left( {\textbf{U}^{1}}^{T} \!\!\!\! \textbf{x}_{t} \right) \! \right) \! \right) \!  \right)  
\end{eqnarray}

\noindent where ${\textbf{W}^{l}}^{T}$ and ${\textbf{U}^{l}}^{T}$  represent the fully connected weight matrices for the recurrent connection and for the $l$-th layer, respectively. 

Because skeleton tracking is an inherently dynamic process, it seems natural to consider DRNNs as a model for supervised learning. As with most researcher, for the first time we train DRNNs, we considered two most popular deep learning techniques, Dropout and Rectified Linear Units (ReLU) \cite{Krizhevsky12}. We used a Rectified Linear Unit (ReLU) as nonlinear activation function for all units in hidden layers. However, unfortunately, dropout does not work well with RNNs unlikely feedforward deep neural networks. Although we carefully applied dropout to DRNNs with our datasets according to the way proposed by \cite{Zaremba12}, we found that dropout leads to divergence. The values of output units are computed by linear activation. 

An alternative for modeling sequences is Long Short-Term Memory (LSTM) \cite{Hochreiter97}. LSTM is a variants of the RNN that perform better on problems with long term dependencies because LSTM has been designed to address the vanishing and exploding gradient problems of conventional RNNs. We trained single layer LSTM and compared performance to single layer RNNs with ReLU activation function for hidden units. In our test dataset, however, LSTM achieved lower performance and took longer time to train. Hence, we did not employ LSTM for supervised learning.

\subsection{Details in Training Two DRNNs}

In the following, we will refer two DRNNs for improving joint position and velocity of skeleton to pDRNN and vDRNN, respectively. pDRNN and vDRNN are five layers, where three layers are hidden and two layers are input and output, respectively. The size of each hidden layer is 256. The number of units in input and output layer 48 because the number of joints to be refined is 16 and each joint is composed of x, y and z coordinates. Kinect v2 supports 25 joints and 16 joints used in our method, which are as follows: \emph{spinebase}, \emph{spinemid}, \emph{neck}, \emph{shoulderleft}, \emph{elbowleft}, \emph{wristleft}, \emph{shoulderright}, \emph{elbowright}, \emph{wristright}, \emph{hipleft}, \emph{kneeleft}, \emph{ankleleft}, \emph{footleft}, \emph{hipright}, \emph{kneeright}, \emph{ankleright}, \emph{footright}, and \emph{spineshoulder}. Among 25 joints, some joints, such as \emph{thumbleft} and \emph{thumbright} are tracked very unstable and some joints are not supported by the motion capture system. \emph{head}, \emph{handleft}, \emph{handright}, \emph{handtipleft}, \emph{thumbleft}, \emph{handtipright}, \emph{thumbright}, \emph{footleft} and \emph{footright} were excluded in our method.    

Temporal lengths of training data for pDRNN is 7. In training phase, absolute joint positions of Kinect skeleton, it is denoted by $\textbf{z}$, is transformed to relative positions with respect to parent joints. The root joint is \emph{spinemid} and only root joint is represented by absolute position. The joints tracked using motion capture system are transformed in the same way. Hence, the output of pDRNN is relative joint positions except \emph{spinemid} joint. The output is transformed to absolute positions and it is denoted by $\tilde{\textbf{z}}$. We do not represent body joint using relative angle because skeleton poses produced by Kinect sensor vary along with the change of the orientation between performer and Kinect sensor. We thus need to preserve angle information in our representation. Temporal lengths of training data for vDRNN is 20. The training data for vDRNN are the velocities of the improved skeleton poses, which is defined by $ \textbf{v}_{t} = \tilde{\textbf{z}}_{t} - \tilde{\textbf{z}}_{t-1}$. We denote the input and output for vDRNN  as $\textbf{v}$ and $\tilde{\textbf{v}}$, respectively. The L-BFGS optimization algorithm is used to train two networks from random initialization and sum-of-squared errors is used for objective functions.

\section{Three Methods for Integrating Improved Position and Velocity of Skeletons}

pDRNN trained based on a large amount of training data can already refine inaccurate Kinect skeleton. However, higher improvement can be expected by integrating pDRNN and vDRNN. In this sense, we propose three methods to combine the outputs produced by pDRNN and vDRNN. First method we have developed is to use K-Nearest Neighbor (KNN). KNN is an instance-based method for classification and regression. In both case, the target value of unknown input is determined according to the values of its $K$ nearest training data. Although the scheme works well it is sensitive to the number of $K$. Thus, we varies the value of $K$ automatically and we will call the variant of KNN as soft-KNN (sKNN) in the following. Second method is based on Kalman filtering. Kalman filter is an algorithm that assumes the true state at time $t$ by observing a series of measurements over time. Specifically, Kalman filter predicts and corrects the estimate based on measurement and process models. The outputs of pDRNN and vDRNN are used for the measurement and process model, respectively. The last method is to combine sKNN and Kalman Filtering. The details will be described in  Section 4.3.      

\subsection{Integrating based on Soft-KNN}

\begin{figure}
 \centering
 \includegraphics[scale=0.5]{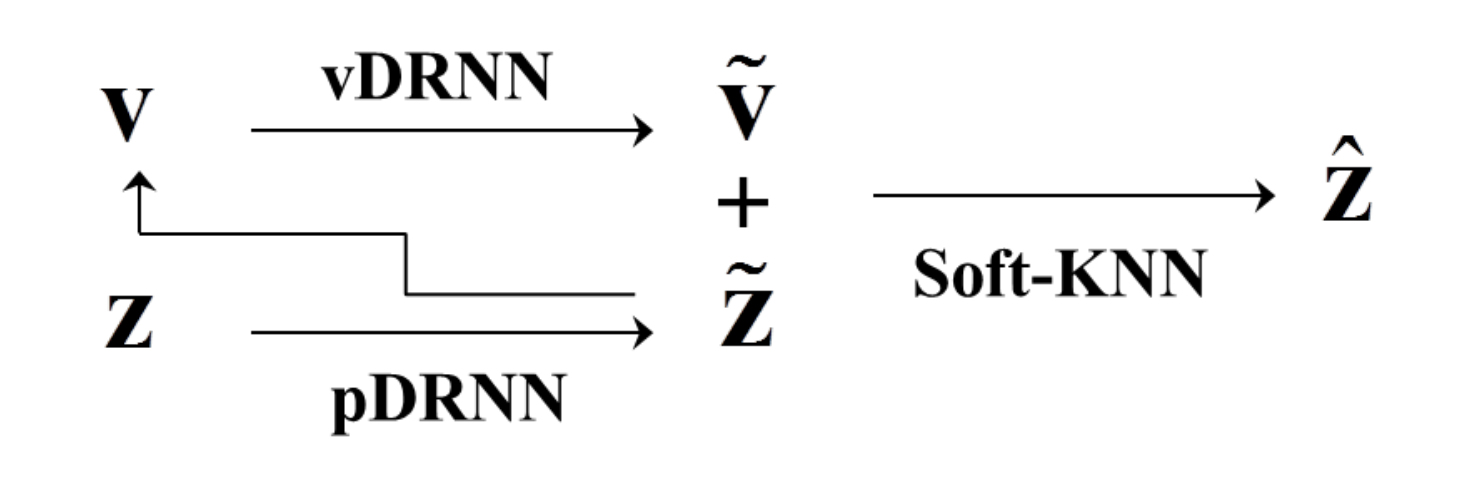} 
 \caption{Schematic representation of soft-KNN.}
\end{figure} 

Figure 1 shows schematic diagram of soft-KNN. Let $S$ = $\lbrace$ ($\tilde{\textbf{z}}_{1}$, $\textbf{z}^{M}_{1}$ ),$\ldots$, ($\tilde{\textbf{z}}_{N}$, $\textbf{z}^{M}_{N}$ ) $\rbrace$ be a set of $N$ input-output training points, where  $\tilde{\textbf{z}}$ is refined skeleton pose by pDRNN and $\textbf{z}^{M}$ is corresponding body joints captured from motion capture system. For a novel pattern $\tilde{\textbf{z}}_{t}$ at time $t$, the proposed soft-KNN regression computes the mean of the target values of its $\tilde{K}$-nearest neighbors. The $j$-th component of the skeleton pose generated by soft-KNN is defined by

\begin{eqnarray}
\label{eq:RNN}
\hat{z}_{t}^{j} = \frac{1}{\tilde{K}} \sum_{ \substack {i \in N_{K} (\tilde{\textbf{z}}_{t}) \\ p_{j} (| v_{i,t}^{j} - \tilde{v}_{t}^{j} |) > \theta } }  z^{j,M}_{i}
\end{eqnarray}

\noindent where set $N_{K} (\tilde{\textbf{z}}_{t})$ contains the indices of $K$-nearest neighbors of $\tilde{\textbf{z}}_{t}$. The number of nearest neighbors for summation is reduced to $\tilde{K}$. $\tilde{K}$ is determined by $p_{j} (| v_{i,t}^{j} - \tilde{v}_{t}^{j} |)$. $\tilde{v}_{t}^{j}$ is the $j$-th component of the velocity generated by vDRNN. $v_{i,t}^{j}$ is velocity of the $j$-th component of the $i$-th training data, which is defined by 

\begin{eqnarray}
\label{eq:RNN}
v_{i,t}^{j} = \tilde{z}_{i}^{j} - \hat{z}_{t-1}^{j}
\end{eqnarray}

\noindent where $\hat{z}_{t-1}^{j}$ is the $j$-th component of the skeleton pose obtained by soft-KNN regression at time $t-1$.  It should be noted that $\hat{z}_{t-1}^{j}$ is used for computing velocity instead of $\tilde{z}_{t-1}^{j}$. This is because $\hat{z}_{t-1}^{j}$ is assumed to be closer to the true joint position than $\tilde{z}_{t-1}^{j}$ it is thus appropriate for calculating current velocity of the $j$-th component of the $i$-th sample. $\hat{z}_{0}^{j}$ is set to $\tilde{z}_{0}^{j}$. We assume that the the initial body pose improved by pDRNN is very close to the skeleton tracked by motion capture system because in our experiment the initial pose of the performer is restricted to standing toward Kinect sensor.  

Two conditions for summation in Equation (4) indicate that if the $j$-th component of velocity of the $i$-th sample is far from the $j$-th component of improved current velocity, although the $i$-th training sample is included in $K$-nearest neighbors, the $j$-th component of the sample is excluded for summation. 
The probability distribution for the $j$-th component in Equation (4) is zero mean Gaussian and is estimated during training phase. Mean and variance are estimated by computing $|v^{j,M}_{i} - \tilde{v}_{i}^{j}|$ on all validation dataset. Here, $v^{j,M}_{i}$ denotes true velocity computed using motion capture data. In this work, $K$ is set to 300 and $\theta$ is 0.05.

\subsection{Integrating based on Kalman Filtering}

In Kalman filter framework, the dynamics and the measurements are modeled by the following discrete-time state-space model:

\begin{eqnarray}
\textbf{x}_{t} = \textbf{F}_{t}\textbf{x}_{t-1} + \textbf{G}_{t}\textbf{v}_{t} + \textbf{w}_{t} \\
\label{eq:State_Measurement model}
\textbf{z}_{t} = \textbf{H}_{t}\textbf{x}_{t} + \textbf{u}_{t}.
\end{eqnarray}

\noindent where \textbf{x}, \textbf{z}, \textbf{v}, \textbf{F}, \textbf{G} and \textbf{H} are the state vector, measurement vector, input control vector, state transition matrix, input transition matrix, and measurement matrix, respectively. It is assumed that \textbf{w} is the process noise vector, which has has zero mean with a covariance matrix \textbf{Q} = $E\{ \textbf{ww}^{T} \}$, and \textbf{u} is the measurement noise vector that also has zero mean with a covariance matrix \textbf{R} = $E\{ \textbf{uu}^{T} \}$. In this work, since we consider an uncorrelated covariance matrix, \textbf{Q} and \textbf{R} become diagonal matrices. In our experiment, \textbf{F}, \textbf{G} and \textbf{H} was set to identity matrix hence prediction model becomes $\textbf{x}_{t} = \textbf{x}_{t-1} + \textbf{v}_{t}$. \textbf{Q} and \textbf{R} were determined by using validation dataset.

The state, $\textbf{x}_{t}$, we should estimate is true skeleton pose and the dimension is 48 as mentioned earlier. Our contribution is to replace the measurement vector, $\textbf{z}_{t}$, with the improved body joints, $\tilde{\textbf{z}}$, and the input control vector, $\textbf{v}_{t}$, with the enhanced velocities, $\tilde{\textbf{v}}_{t}$.  Therefore, the $j$-th row and $j$-th column of $\textbf{R}$ and $\textbf{Q}$ are determined by computing $(z^{j,M}_{i} - \tilde{z}_{i}^{j})$ and $(v^{j,M}_{i} - \tilde{v}_{i}^{j})$, respectively. In our methods, $\textbf{x}_{0}$ was set to $\tilde{\textbf{z}}_{0}$.

\subsection{Integrating based on combination of Soft-KNN and Kalman Filtering}

\begin{figure}
 \centering
 \includegraphics[scale=0.45]{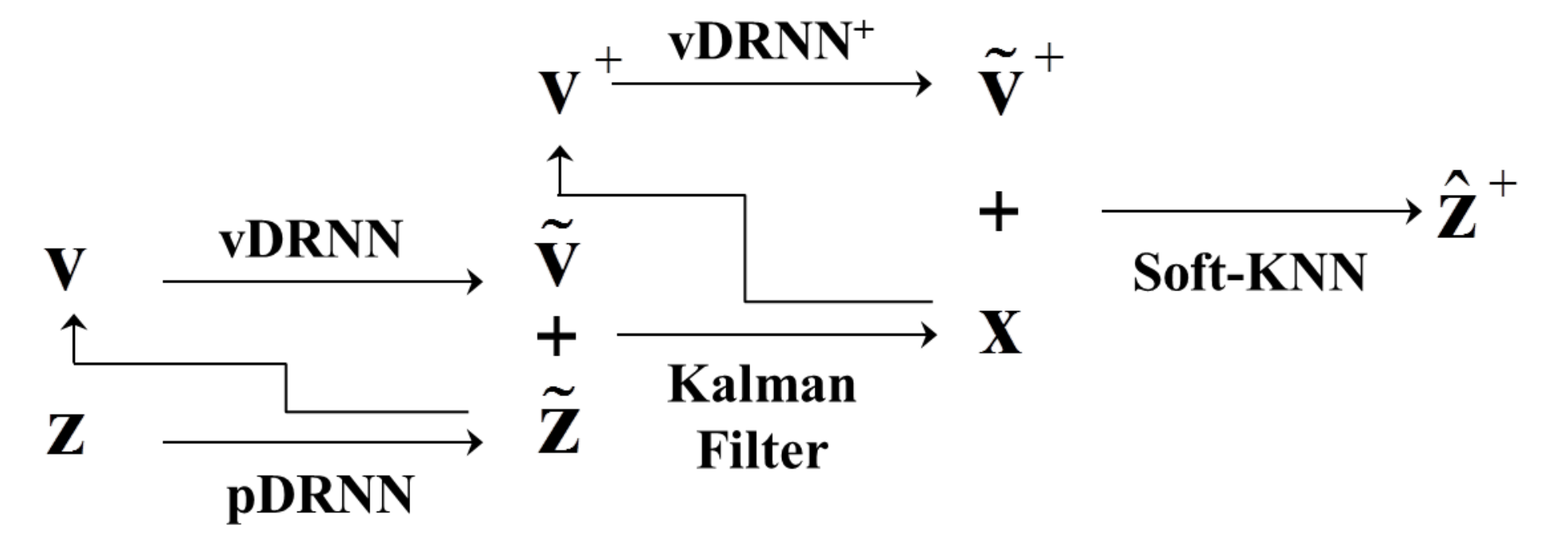} 
 \caption{Schematic representation of sKNNkF.}
\end{figure}

The last method is to combine soft-KNN and Kalman Filtering methods described above. We will refer the method to sKNNkF in the following. Figure 2 shows schematic diagram of sKNNkF. Let $S^{+}$ = $\lbrace$ ($\textbf{x}_{1}$, $\textbf{x}^{M}_{1}$ ),$\ldots$, ($\textbf{x}_{N}$, $\textbf{x}^{M}_{N}$ ) $\rbrace$ be a set of $N$ input-output training points, where  $\textbf{x}$ is estimated by Kalman filtering and $\textbf{x}^{M}$ is corresponding skeleton pose captured from motion capture system. For a novel pattern $\textbf{x}_{t}$ at time $t$, the soft-KNN regression computes the mean of the target values of its $\tilde{K}$-nearest neighbors. The $j$-th component of the skeleton pose generated by soft-KNN is defined by

\begin{eqnarray}
\label{eq:RNN}
\hat{z}_{t}^{+,j} = \frac{1}{\tilde{K}} \sum_{ \substack {i \in N_{K} (\textbf{x}_{t}) \\ p_{j}^{+} (| v_{i,t}^{+,j} - \tilde{v}_{t}^{+,j} |) > \theta^{+} } }  x^{j,M}_{i}
\end{eqnarray}
  
\noindent Here, $v_{i,t}^{+,j}$ is velocity of the $j$-th component of the $i$-th training data, which is defined by 

\begin{eqnarray}
\label{eq:RNN}
v_{i,t}^{+,j} = x_{i}^{j} - \hat{z}_{t-1}^{+,j}
\end{eqnarray}

The probability distribution for the $j$-th component in Equation (8) is zero mean Gaussian and is estimated during training phase. The mean and variance are estimated by computing $|v^{j,M}_{i} - \tilde{v}_{i}^{+,j}|$ on all validation dataset. In this work, $K$ is set to 300 and $\theta^{+}$ is 0.05. Here, $\tilde{v}_{i}^{+,j}$ denotes the improved velocity obtained using another deep recurrent neural network. We call the network as vDRNN$^{+}$. Input training data for vDRNN$^{+}$ is velocity of estimated skeleton pose in Kalman filtering step, which is defined by $ \textbf{v}_{t}^{+} = \textbf{x}_{t} - \textbf{x}_{t-1}$. We denote the output for vDRNN$^{+}$ as $\tilde{\textbf{v}}^{+}$. The network has identical structure with vDRNN and the temporal length of training data is also same.

\section{A Novel Measure for Evaluating Human-like Natural Movement}

As mentioned earlier, our goal is to propose a skeleton tracking method, in which captured body joint trajectories should be human-like natural movement. Most popular measure to evaluate quality of tracked skeleton pose is average position error (APE). If APE of a sequence of 3D positions is less than 1mm, the estimated trajectory can be considered as human-like movement. In fact, this condition extremely difficult to meet. However, we found that if APEs of two skeleton trajectories are 3cm and 4cm, respectively, in that case we cannot be confident that which is better movement. Suppose that two tracked trajectories. The former is a joint trajectory that has a large number of small vibrated motions. In contrast, the latter trajectory consists of natural movements but the orientation of the tracked body center is little bit different to that of the ground truth body center. In this case, APE of the latter is often larger than that of the former. Therefore, an investigation for a novel measure to assess human-like natural movement is required.       

Flash and Hogan have proposed that the human motor system minimizes jerk \cite{Flash85}. Jerk is the 3rd derivative of the position trajectory. In this sense, some researchers have developed human motion prediction techniques based on the minimum jerk model \cite{Thobbi11}, \cite{Corteville07}. However, the minimum jerk model assumption fails if the human decides to change the course of the trajectory during performing activity. We also found jerks of some actions such as, kicking or punching are not low. Hence, we define jerk error (JE) of $j$-th component of tracked skeleton at time $t$ as

\begin{eqnarray}
\label{eq:Jerk}
JE = |j_{t}^{j} - j_{t}^{j,M}|
\end{eqnarray}

\noindent where $j_{t}^{j,M}$ is jerk of the trajectory captured by motion capture system. We argue that average jerk error (AJE) can evaluate naturalness of captured motions in terms of vibrated and discontinuous motions. However, AJE only cannot evaluate the quality of tracking appropriately. Suppose that an extreme case. If one activity is standing and the other is sitting the jerks of two activities are identical. Hence, in our experiment, we consider APE as well as AJE.

\begin{figure*}
 \centering
 \includegraphics[scale=0.6]{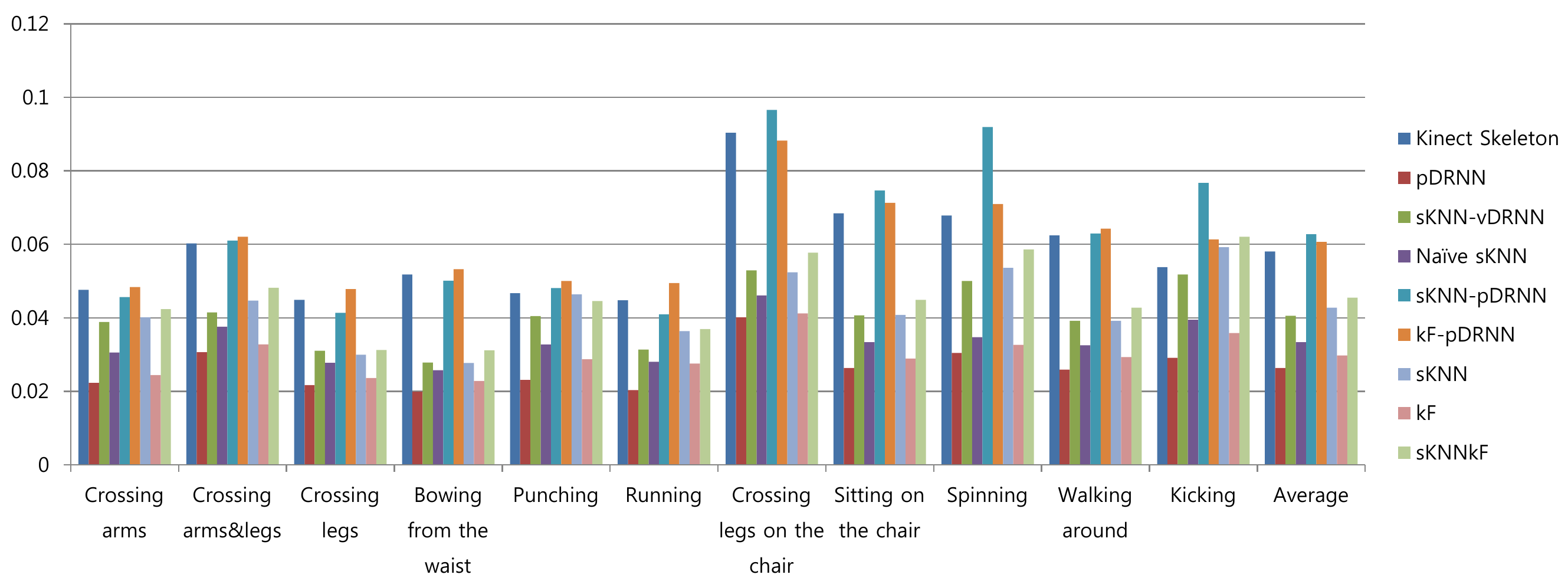} 
 \caption{The average position error (APE).}
\end{figure*} 

\begin{figure*}
 \centering
 \includegraphics[scale=0.6]{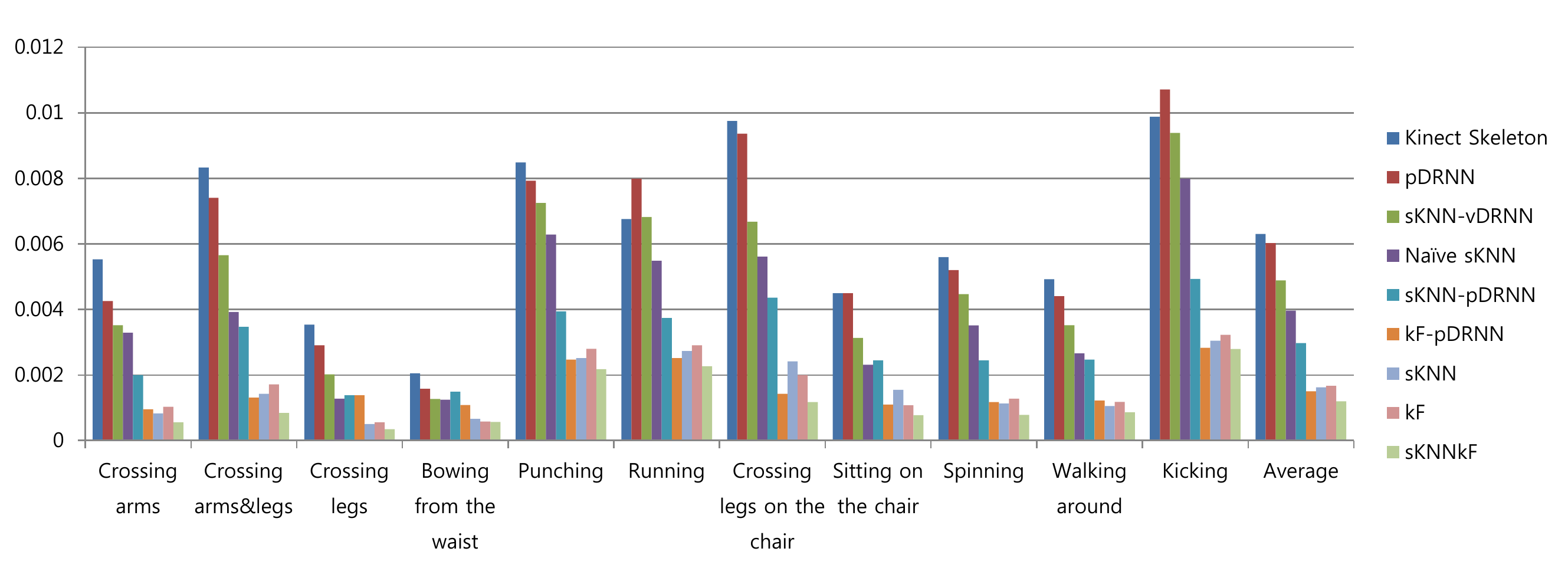} 
 \caption{The average jerk error (AJE).}
\end{figure*}

\section{Experiments}

\subsection{Experimental Setup}

We implemented the algorithm proposed in this paper using MATLAB and the Microsoft Kinect SDK 2.0 on Window 8 OS. All experimental tests were run on a PC with an Intel Core i5 1.8GHz processor and 4GB RAM. The Microsoft Kinect SDK 2.0 can extract skeleton data at approximately 30 frames per second (fps). For supervised learning and evaluation, we employed an OptiTrack motion capture system to provide a set of ground truth trajectories. Kinect sensor and motion capture system tracked skeleton poses simultaneously with recoding capturing time hence we can construct sets of input and target data pairs. Kinect sensor and the motion capture system extrinsically calibrated using least-squares solution.

We collected training, validation and test dataset. The training and validation dataset is composed of free movements human can do. Validation dataset was employed to decide structure of DRNNs such as the number of layers, the number of hidden neuron size and the temporal length of training data. The variances of Gaussian distributions used in soft-KNN and the covariance matrices \textbf{R} and \textbf{Q} used in Kalman filtering were also determined using Validation dataset. The numbers of frames in training and validation dataset are 45,179 and 6,483. 
As mentioned earlier, the temporal lengths of training data for pDRNN, vDRNN and vDRNN$^{+}$ is 7, 20 and 20, respectively. To construct dataset to train deep recurrent neural networks, we sampled sets of sequence of data with a temporal stride 1.

Test dataset consists of 11 types of activity classes such as \emph{Crossing arms}, \emph{Crossing arms and legs}, \emph{Crossing legs}, \emph{Bowing from the waist}, \emph{Punching}, \emph{Running}, \emph{Crossing legs on the chair}, \emph{Sitting on the chair}, \emph{spinning}, \emph{walking around} and \emph{kicking}. Some activities such as, \emph{Crossing arms and legs}, \emph{Sitting on the chair} consist of a large amount of severe self-occlusion poses while \emph{Running} and \emph{Bowing from the waist} include a small number of self-occlusion poses. Each activity class was repeated ten times. Every activity start with standing pose and then repeat a certain activities such as, \emph{Crossing arms} \emph{Punching} several times. An activity is composed of approximately 150$\sim$250 frames. The total numbers of frames in test dataset is 20,508. 

Every activity except \emph{Spinning} and \emph{Walking around} were performed facing the Kinect sensors. For the cases of \emph{Spinning} and \emph{Walking around}, the minimum and maximum orientations relative to the Kinect sensor were -90$^{\circ}$ and 90$^{\circ}$, respectively, We did not allow the Kinect sensor to look at the performer's back because Kinect skeleton tracker cannot distinguish front and back. The average distance from the Kinect sensor to the human was about 3m and the height of Kinect above the ground plane was 130cm.

\subsection{Experimental Results}

We have implemented three skeleton tracking techniques: (1) sKNN(Integrating pDRNN and vDRNN based on soft-KNN), (2) kF(Integrating pDRNN and vDRNN in Kalman filter framework) and (3)sKNNfF(Integrating pDRNN, vDRNN and vDRNN$_{+}$ based on combination of soft-KNN and Kalman filtering). We have additionally implemented six skeleton tracking techniques for the sake of comparison: (1) Kinect Skeleton (2) pDRNN (Skeleton tracking using pDRNN), (3) sKNN-pDRNN (sKNN without pDRNN), (4) sKNN-vDRNN (sKNN without vDRNN), (5) n\"aive-sKNN (sKNN with using $\tilde{z}_{t-1}^{j}$ instead of $\hat{z}_{t-1}^{j}$ in Equation (5)) and (6) kF-pDRNN (kF without pDRNN). sKNN-pDRNN choose $K$-nearest neighbors of $\textbf{z}_{t}$ instead of $\tilde{\textbf{z}}_{t}$ and a dataset $S^{-}$ = $\lbrace$ ($\textbf{z}_{1}$, $\textbf{z}^{M}_{1}$ ),$\ldots$, ($\textbf{z}_{N}$, $\textbf{z}^{M}_{N}$ ) $\rbrace$ is used for training. sKNN-vDRNN reduces the number of nearest neighbors from $K$ to $\tilde{K}$ using $p_{j} (| v_{i,t}^{j} - v_{t}^{j} |)$ instead of $p_{j} (| v_{i,t}^{j} - \tilde{v}_{t}^{j} |)$. $p_{j} (| v_{i,t}^{j} - v_{t}^{j} |)$ is estimated by computing $|v^{j,M}_{i} - v_{i}^{j}|$. kF-pDRNN employs $\textbf{v}_{t}$ for the control input and $\textbf{Q}$ are determined by computing $(v^{j,M}_{i} - v_{i}^{j})$. We did not implement kF-vDRNN since kF-vDRNN produces identical results to pDRNN. In kF-vDRNN, $\textbf{v}_{t}=\textbf{z}_{t}-\textbf{z}_{t-1}$ and the prediction model becomes $\textbf{x}_{t} = \textbf{x}_{t-1} + \textbf{v}_{t}$. And if $\textbf{x}_{0}$ is set to $\textbf{z}_{0}$, $\textbf{x}_{1}$ becomes equal to $\textbf{z}_{1}$. In this way, $\textbf{x}_{t} = \textbf{z}_{t}$ for all time $t$.

\begin{figure*}
 \centering
 \includegraphics[scale=0.45]{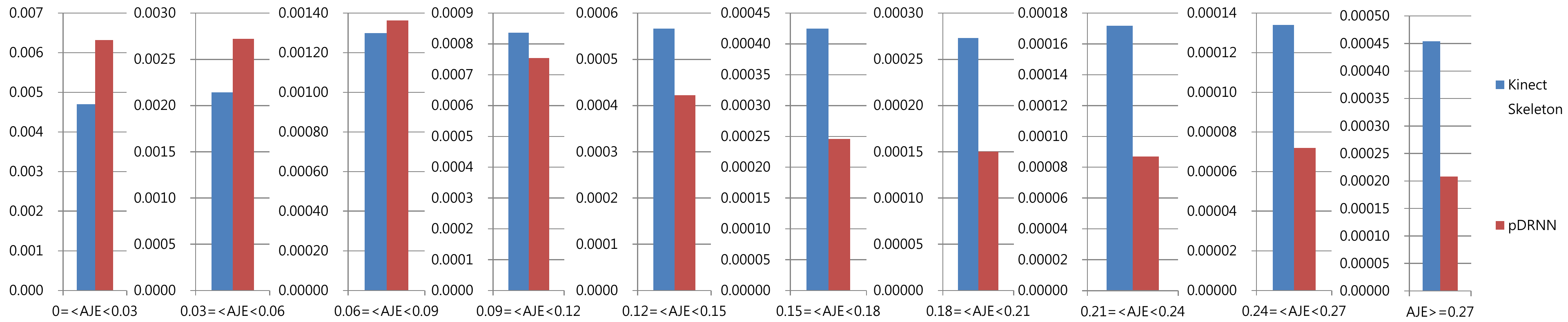} 
 \caption{The average jerk error histogram for Kinect skeleton and pDRNN.}
\end{figure*} 

\begin{figure*}
 \centering
 \includegraphics[scale=0.45]{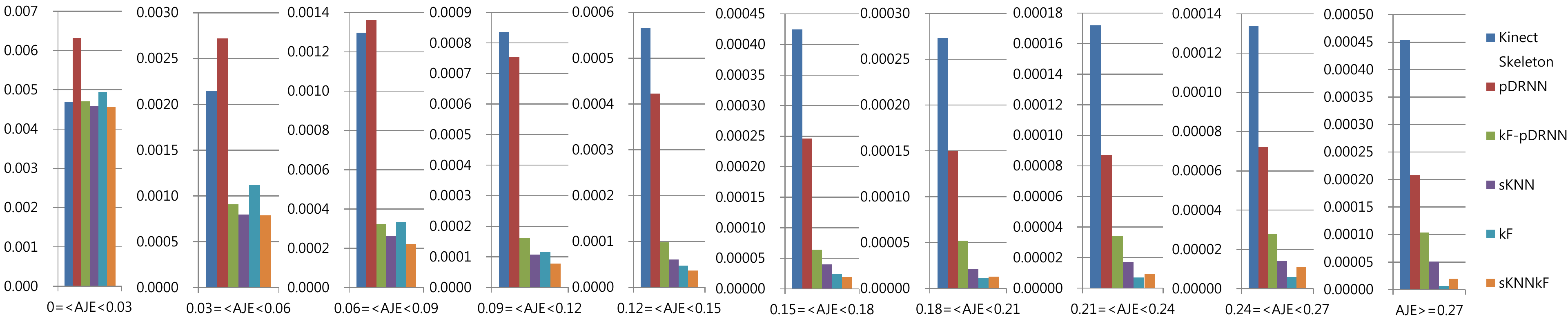} 
 \caption{The average jerk error histogram for Kinect skeleton, pDRNN, kF-pDRNN, sKNN, kF and sKKkF.}
\end{figure*}

\begin{figure*}
 \centering
 \includegraphics[scale=0.55]{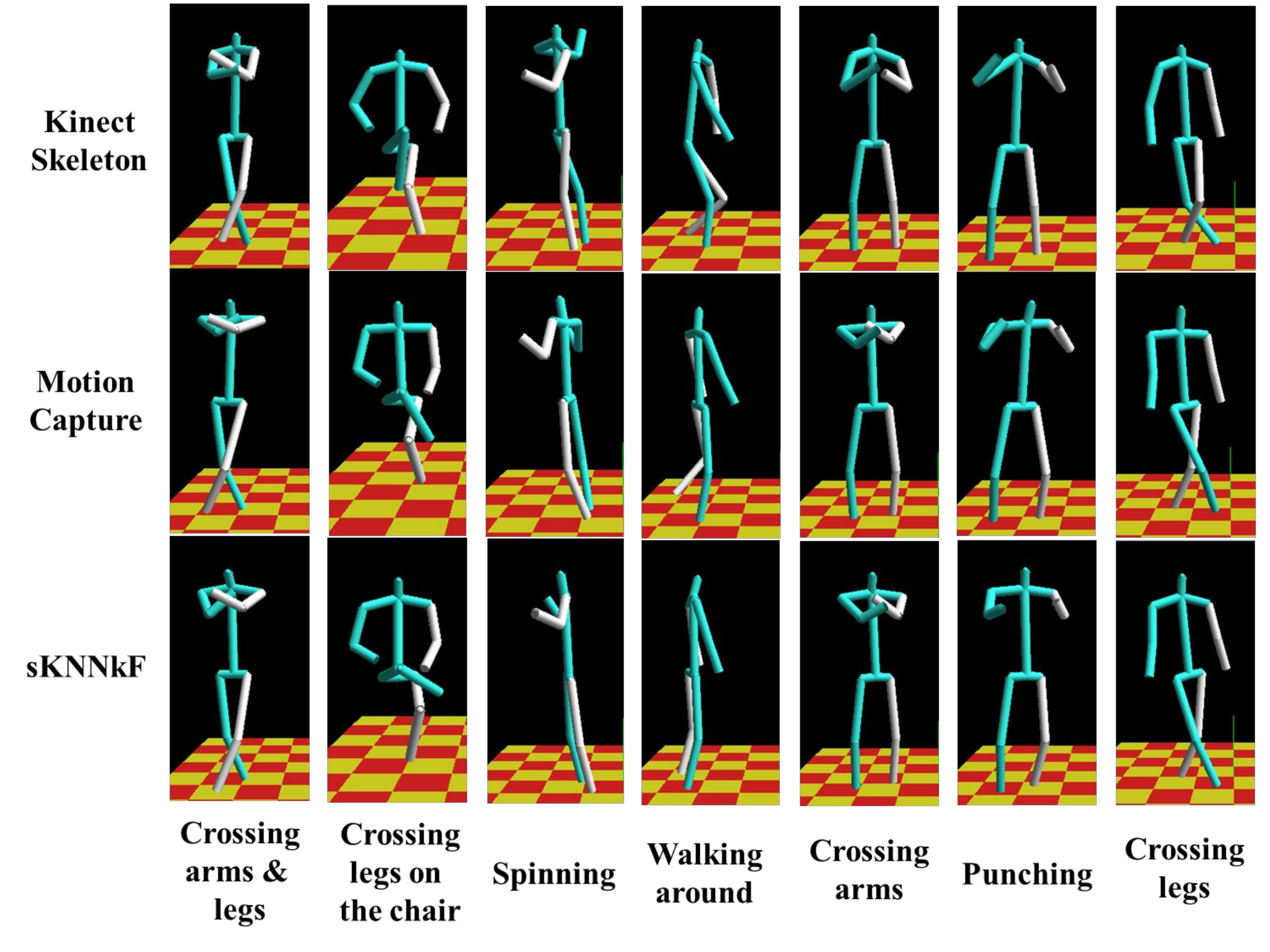} 
 \caption{The skeleton poses tracked by Kinect skeleton tracker, motion capture system and sKKkF.}
\end{figure*}

Figure 3 shows average position error (APE). APE of Kienct skeleton is 0.058 and pDRNN decrease APE to 0.026. pDRNN achieves considerable reduce. APEs of kF, sKNN and sKNNkF are 0.0297, 0.0427 and 0.0454, respectively. There is small increase in APE of kF compared to APE of pDRNN. In contrast, APE of sKNN is relatively larger than that of kF. It seems because KNN estimates current pose depend on simple combination of nearest training samples. sKNNkF shows a little bit worse performance than KNN and APE of sKNNkF is highest among three proposed methods. It is observed that APE is accumulated through sKNN and kF. In cases of sKNN-pDRNN and kF-pDRNN, the APEs are even higher than APE of Kinect skeleton. We can conclude that pDRNN plays a important role to reduce APE and additional procedures after the regression using pDRNN increase APE.

Figure 4 shows average jerk error (AJE). It is noted that AJEs of sKNN, kF and sKKkF achieve best performance (0.0016, 0.0016 and 0.0011, respectively). kF-pDRNN performs similar AJE to the proposed three methods, but as shown in Figure 3 APE of kF-pDRNN worse than that of Kienct skeleton. We can conclude that vDRNN plays a important role to reduce AJE. However, although n\"aive-sKNN integrates both pDRNN and vDRNN the reduction of AJE is small.

According to Figure 4, AJEs of Kinect skeleton and pDRNN are similar, which are 0.0063 and 0.006, respectively. It seems that pDRNN has a little influence on improving vibrated and discontinuous movement. To investigate effects of pDRNN on tracking human-like natural motion, we construct average jerk error histogram. First, we divide the entire range of jerk error into ten bins and then AJE for each bin is computed. Specifically, for example, last bin summates a jerk error greater than 0.27 and the summation is divide by $M$. $M$ = (the dimension of skeleton pose) * (the number of total frame in test dataset). It should be considered that jerk errors fall into first bin is usually very small vibrated motions thus it cannot be hard to recognize.  

The AJE histogram for Kinect skeleton and pDRNN is shown in Figure 5. It is observed that AJE of pDRNN is larger than Kinect skeleton in first three bin, while pDRNN achieves smaller error than Kinect skeleton in the rest. This implies that pDRNN reduces moderate and large jerk errors. In our experiments, we found that pDRNN cannot remove small vibrated movement but are effective to alleviate severe discontinuity.

Figure 6 shows an average jerk error histogram for Kinect skeleton, pDRNN, kF-pDRNN, sKNN, kF and sKNNkF. It is observed that sKNN shows better performance than kF from 1 to 4 bins while kF shows better performance than sKKN from 5 to 10 bins. sKNNkF achieves good performance in the entire range of histogram. kF-pDRNN seems not to be effective to reduce moderate and large jerk errors. The proposed three methods can reduce the unnatural movements range from small vibration to severe discontinuity.

For qualitative comparison, the tracking results produced by Kinect skeleton tracker, motion capture system and sKKkF are displayed in Figure 7. The seven images in each row were chosen during \emph{Crossing arms and legs}, \emph{Crossing legs on the chair}, \emph{Spinning},  \emph{walking around}, \emph{Crossing arms}, \emph{Punching} and \emph{Crossing legs} behaviors, respectively. The three images in each column were generated by Kinect skeleton tracker, motion capture system and sKKkF, respectively. It is observed that significant error was produced by the Kinect skeleton tracker. The pose chosen during \emph{Crossing legs on the chair} activity dose not cross legs and the poses selected during \emph{Spinning} and \emph{walking around} activities are quite different to ground truth, whereas the skeletons generated by sKKkF look similar to the ground truths and seem to reflect the natural movement of the performer.

\section{Conclusions}

The goal of this paper was to propose a method to improve Kinect skeleton, vibrated and discontinuous when self-occlusion occurs, to human-like natural motion. To this end, we first employed deep recurrent neural networks to refine the position and velocity errors of the skeleton poses. Then, we proposed three methods to integrate enhanced joint positions and velocities. Moreover, we suggested a novel measure to evaluate naturalness of captured motions. 

We evaluated the proposed methods by comparison with the ground truth acquired from a commercial motion capture system and compared the results to those of Kinect skeleton and of several variants of our methods. Our proposed three approaches performed considerably better than the Kinect skeleton tracker and the proposed integration methods leads further improvement than when we refine Kinect skeleton data using only pDRNN.

\end{document}